\documentclass[letterpaper, 10 pt, conference]{ieeeconf}  

\IEEEoverridecommandlockouts                              

\usepackage{graphicx} 
\usepackage{listings}
\usepackage{spverbatim}
\usepackage{fancyvrb}
\usepackage{lipsum} 
\usepackage[utf8]{inputenc}
\usepackage{array}
\usepackage{xcolor}
\usepackage{hyperref}

\title{\LARGE \bf
Smart Cloud: Scalable Cloud Robotic Architecture for Web-powered Multi-Robot Applications
}

\author{Manoj Penmetcha, Shyam Sundar Kannan, and Byung-Cheol Min
\thanks{The authors are with the SMART Lab, Department of Computer and Information Technology, Purdue University, West Lafayette, IN 47907, USA
	{\tt\small mpenmetc@purdue.edu | kannan9@purdue.edu | minb@purdue.edu}}%
}

\begin{document}
\maketitle
\thispagestyle{empty}
\pagestyle{empty}

\begin{abstract}
 Robots have inherently limited onboard processing, storage, and power capabilities. Cloud computing resources have the potential to provide significant advantages for robots in many applications. However, to make use of these resources, frameworks must be developed that facilitate robot interactions with cloud services. In this paper, we propose a cloud-based architecture called \textit{Smart Cloud} that intends to overcome the physical limitations of single- or multi-robot systems through massively parallel computation, provided on demand by cloud services. Smart Cloud is implemented on Amazon Web Services (AWS) and available for robots running on the Robot Operating System (ROS) and on the non-ROS systems. \textit{Smart Cloud} features a first-of-its-kind architecture that incorporates JavaScript-based libraries to run various robotic applications related to machine learning and other methods. This paper presents the architecture and its performance in terms of CPU usage and latency, and finally validates it for navigation and machine learning applications.
\end{abstract}


\section{Introduction}
The scope of the robotics industry is immense, and the industry is poised to see huge gains in the coming years. The International Data Corporation estimates the 2019 economic value of robotics and related services will hover around \$135.4 billion, and during the period of 2018 to 2023, the industry is estimated to register a compound annual growth of 24.52 percent \cite{CloudRob32:online}. The ubiquitous availability of big data and recent advancements in machine learning can be used to develop smarter and more responsive robots. Most such applications involve processing large quantities of data, which requires high-performing computational resources \cite{7006734, 8913967}. However, existing robots come with limited onboard computing capabilities, and once a robot is built, it is not easy to change the hardware configuration. By enabling cloud computing for robotic applications, robots will be able to access increased computational power and storage space as needed to carry out their assigned tasks. With the resources provided by cloud computing services, computationally-intense robotic tasks like object detection, navigation, and others can be solved more efficiently.

Cloud computing is a service-driven paradigm for hosting applications on remote infrastructure, i.e. resources are provided on-demand. Since its inception, cloud computing has helped researchers and business users to host applications by providing access to distributed and shared computing resources over the Internet. In practice, the services provided by the cloud can be categorized into three major types: Software as a Service (SaaS), Platform as a Service (PaaS), and Infrastructure as a Service (IaaS) \cite{5071529}. 

SaaS is used to provide access to a completely developed application over the Internet. IaaS refers to the on-demand provisioning of infrastructural resources. In IaaS, the consumer is usually provided with fundamental computing resources, where the consumer can deploy an arbitrary software. PaaS falls between IaaS and SaaS, where cloud service providers provide the consumers with tools like a programming languages, libraries and so on, to help the consumers to develop the applications.

Cloud robotics, first introduced as a term by James J. Kuffner in 2010 \cite{10031099795}, can be defined as the wireless connection of robots to external computing resources to support robot operation. Cloud-enabled robots are able to offload computing tasks to remote servers, thus relying less on their onboard computers and instead exploiting the inexpensive computing power and data storage options provided by cloud service providers. The cloud robotics market is estimated to achieve 23.2 percent compound annual growth. The cloud robotics market was valued around \$2.3 billion  in 2017, and is expected to reach \$7.9 billion by the year 2023 \cite{CloudRob32:online}. To realize the full potential and scope of cloud robotics, however, it is very important to innovate and address the shortcomings currently faced by the field.

\begin{figure*}[!t]
\vspace{2mm}
\centering
\includegraphics[scale=0.7
]{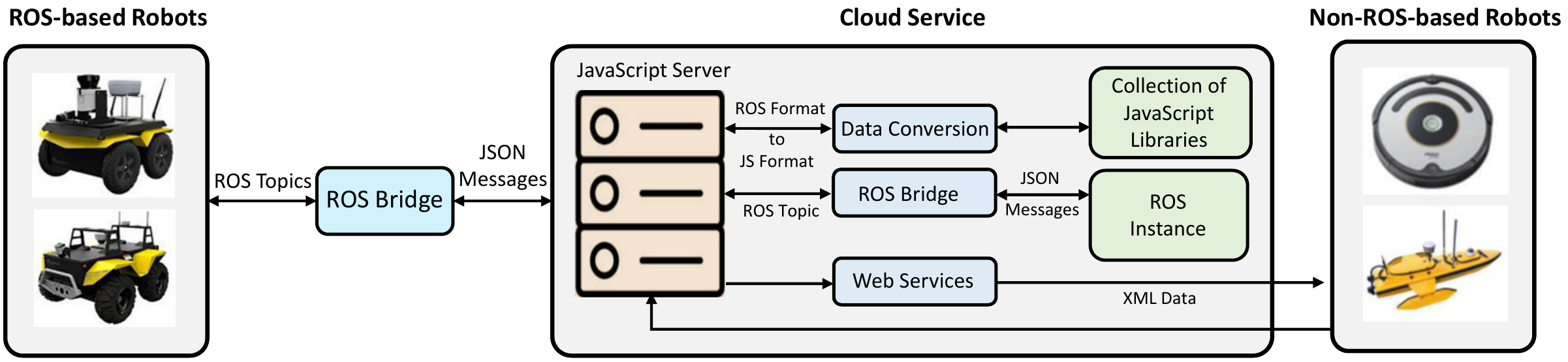}
\caption{Smart Cloud architecture; (Middle) Cloud Server (Cloud Layer): The JS server interacts with JS libraries and ROS, (Left) ROS-Based Robots (Robot Layer with ROS robots): Robots interact with the server using the Rosbridge protocol, and (Right) Non-ROS-Based Robots (Robot Layer with non-ROS robots): Robots send information through wireless data transfer and receive responses as web services.}
\label{fig:Smart Cloud Architecture}
\end{figure*}

In this paper, we present a new cloud robotic architecture called \textit{Smart Cloud}, outlined in Fig. \ref{fig:Smart Cloud Architecture}. The architecture includes several novel functionalities that make it the first of its kind. Smart Cloud can be used as both SaaS and PaaS depending on the needs of the application. In a SaaS-based approach, it provides a simple web-based interface by which a robot can make use of several ready-to-use applications, both Robot Operating System (ROS)-based and non-ROS-based. The architecture backend is built on a JavaScript (JS) server. Using JS inherently allows access to open-source libraries, including machine learning libraries like TensorFlow, data compression mechanisms to reduce network load, and more \cite{10.5555/551216}. Additionally, we intend to provide the developed framework as an open-source application for other researchers to modify, as researchers frequently need the flexibility to develop or customize applications to suit their specific requirements. By making the framework open-source, the proposed architecture can also be used in a PaaS implementation.

In this paper, the functionality of the architecture is demonstrated for several application scenarios. Furthermore, the architecture is evaluated in terms of CPU utilization, and latency.
The remainder of the paper is organized as follows: Section \ref{sec:rel_work} presents a literature review, while Section \ref{sec:architecture} describes the architecture in detail. Section \ref{sec:application} discusses its applications and the evaluation of its performance. Finally, Section \ref{sec:conclusion} presents conclusions and future directions.

\section{Related Work}
\label{sec:rel_work}
The invention of the World Wide Web in the late 1980’s opened up the possibility of connecting a robot to an external machine over the Internet. In 1994, Ken Goldberg was amongst the first to successfully connect a robot to the web, teleoperating the robot through an Internet browser \cite{525358}. In 1997, Masayuki Inaba at the University of Tokyo worked on a so-called “remote brain”; in the associated publication, the authors described advantages of using remote computing for robots \cite{doi:10.1177/02783640022067878}. In 2009, the ‘RoboEarth’ project was announced with the intent to develop a World Wide Web equivalent for robots \cite{5876227}. RoboEarth was developed by a team of researchers from Eindhoven University; the idea was to build a large database that allowed robots to share learned information with one another. The RoboEarth project team created a platform for cloud computing, named Rapyuta \cite{6853392}. Rapyuta is a PaaS-based robotics framework that allows robots to offload all computation-intense tasks to the cloud service. This framework has access to the RoboEarth data repository, which enables robot access to all extant libraries on RoboEarth.

Interest in cloud robotics has led to the development of new vertical research involving architectures that facilitate robot communication with cloud service providers. One of the first architectures in this area was ‘DaVinci’, which used cloud computing infrastructure to generate 3-D models for robot localization and mapping much faster than possible using on-board hardware \cite{5509469}.

Doriya et al. proposed a robot cloud framework that helps low-cost robots offload computationally-intense tasks to the cloud \cite{6398208}. The central unit of the framework is equipped with a ROS master node that facilitates all communication. In 2016, Wang et al. proposed a hybrid frame work called RoboCloud \cite{Li2018}. RoboCloud differs from other cloud robotics architectures by introducing a task-specified mission cloud with controllable resources and predictable resources. For tasks beyond the capability of the mission cloud, the framework opts to utilize public clouds, the same as any other cloud robotics architecture.

In addition to the above-mentioned PaaS  architectures, re searchers have also worked on SaaS-based frameworks, also called Robot as a Service (RaaS). Notably, SaaS-based architectures can more easily overcome interoperability issues that arise due to differing robot hardware. Frameworks such as C2tam \cite{RIAZUELO2014401} and XBotCloud \cite{inproceedings121} focus on specific applications or algorithms, for example object recognition or object grasping. Tian et al. \cite{7989192} proposed a RaaS-based robotics model called Brass, which allows robots to access a remote server that hosts a grasp planning technique. The framework leverages Docker to allow for implementing algorithms by writing simple wrappers around existing code.

In summary, most currently-available architectures are geared towards PaaS models; few of them take a SaaS approach. These architectures are designed to offload robotic applications to the cloud infrastructure. The proposed framework differs from extant frameworks by being built for use in both SaaS-based and PaaS-based applications. A simple web interface will be provided for SaaS usage, and the code will be made publically available for users to develop applications on top of it in PaaS usage. Furthermore, the proposed framework is the first of its kind to provide access to open-source JS libraries; users are not limited to available robotic libraries and do not need to develop applications from scratch, but can make use of available JS libraries for robotic applications. 

\section{Architecture}
\label{sec:architecture}

As shown in Fig. \ref{fig:Smart Cloud Architecture}, the Smart Cloud architecture consists of two main components: the Robot layer (ROS-based robots and non-ROS-based robots) and the Cloud Service layer. We chose JS for the development of this framework because of its ubiquitous nature, support for ROS, and the vast availability of libraries. In terms of available open-source libraries, JS outnumbers every other programming language \cite{10.5555/551216}. Additionally, the JS library Roslibjs allows interaction with the ROS interface through Rosbridge, which is developed for non-ROS users and used to send and receive data in the form of JavaScript Object Notation (JSON) packets. Roslibjs supports essential ROS functionalities such as publishing and subscribing to topics, services, actionlib, and more. 

\subsection{Robot Layer}
The robot layer consists of a single robot or a multi-robot system that runs on either ROS or any generic robot software. In the next subsection, we discuss how the architecture handles data based on the robot’s software and the type of application service requested.

\subsubsection{ROS Based Robots}
ROS is a framework for writing robotic software. The Rosbridge package allows a ROS-based robot to interact with any non-ROS system \cite{7354021} through web sockets. In this architecture, we use Rosbridge to establish a communication protocol between robots and cloud services. On the cloud side of the architecture, a master instance of ROS helps the robots to offload computationally intense tasks to the cloud.

Through this framework, robots can access ROS packages and JS libraries; however, to use ROS packages on the cloud, the robots themselves need to run on ROS. Data from the robot is published as ROS topics and received on the cloud side using Rosbridge with the following code: \\

\vspace{-5pt}
\begin{Verbatim}[fontsize=\footnotesize, frame=single]
function rosTopics()
{
    var topicsClient = new ROSLIB.Service({
    ros : ros,
    name : '/rosapi/topics',
    serviceType : 'rosapi/Topics'
    });
}
\end{Verbatim}

\begin{figure}
\vspace{2mm}
\centering
\includegraphics[width=0.42\textwidth]{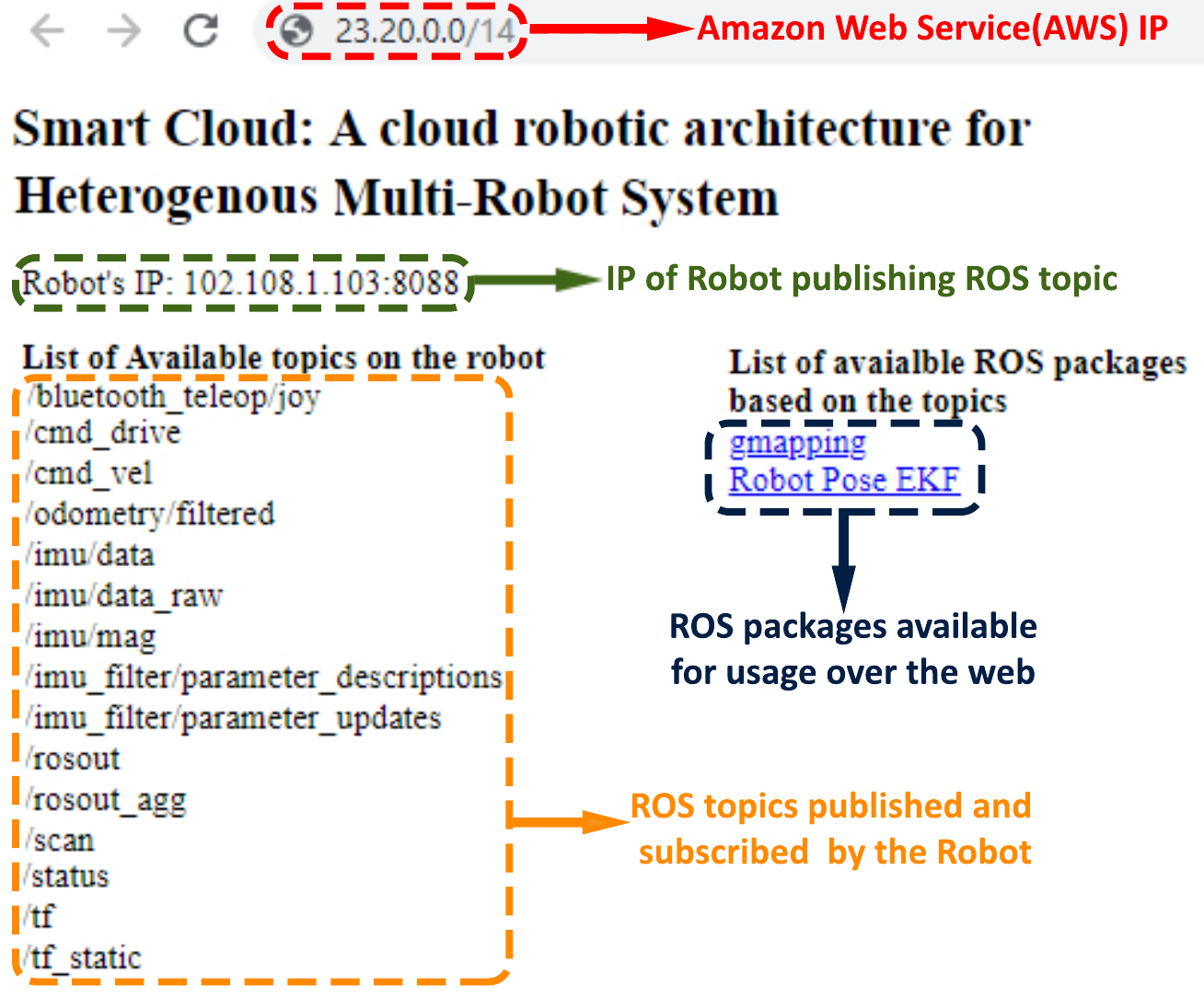}
\vspace{-5pt}
\caption{Smart Cloud: SaaS-based ROS interface. The robot provides a list of topics and the architecture displays corresponding packages that can be used.}
\label{fig:WebPage}
\end{figure}

Almost every ROS package takes topics from the robot as input. Once the cloud framework receives this list of topics, it parses through the list of available ROS packages to find those that can be used with the given input. The list of matching ROS packages are displayed on the web interface for the user to choose between. After the user picks a package, the result is computed and sent back to the robot over the Rosbridge. For example, the ROS gmapping package requires \textit{tf} and \textit{scan} topics. If the framework finds \textit{tf} and \textit{scan} topics available, the user will be provided with the option to use gmapping. Fig. \ref{fig:WebPage}  shows the web interface displayed to a user.

\subsubsection{Non-ROS Based Robots}

The data from robots not using ROS will be transferred directly to the cloud using wireless protocols. Once the data is received, the architecture provides a set of JS-based libraries to choose from based on the message type. For example, if the message is in the form of an image, the framework provides libraries related to object detection, object tracking, and the like. If the message is in the format of GPS coordinates, the framework provides various GPS-based applications.

On the server side, the framework implements a RESTful-based web service that is used to communicate the results back to the robot. Web service is a consistent medium for communication between the client and the server over the internet. The communication between the client and the server is carried out through Extensible Markup Language (XML). In our architecture, we treat the robots as a client and the architecture as a server. The robot sends a HTTP request to the architecture and the architecture sends the response in the form of an XML.

\subsection{Cloud Layer}
Robots are connected to the JS server on the remote cloud infrastructure through web sockets. Depending on application requirements, the cloud layer runs a single or multiple instances of Linux-based operating systems. On these instances, JS Server and ROS are deployed. In this section, we will look into how data is received and processed by the JS server and how various libraries are used for robotic applications.
\subsubsection{JavaScript Server}
In this architecture, we use a JS server based on Node.js \cite{5617064}. Node.js  is an open-source and cross-platform JS server that runs JS scripts outside a browser. The framework has different mechanisms for handling data from ROS and non-ROS systems.

\begin{figure}[!b]
\includegraphics[width=\columnwidth]{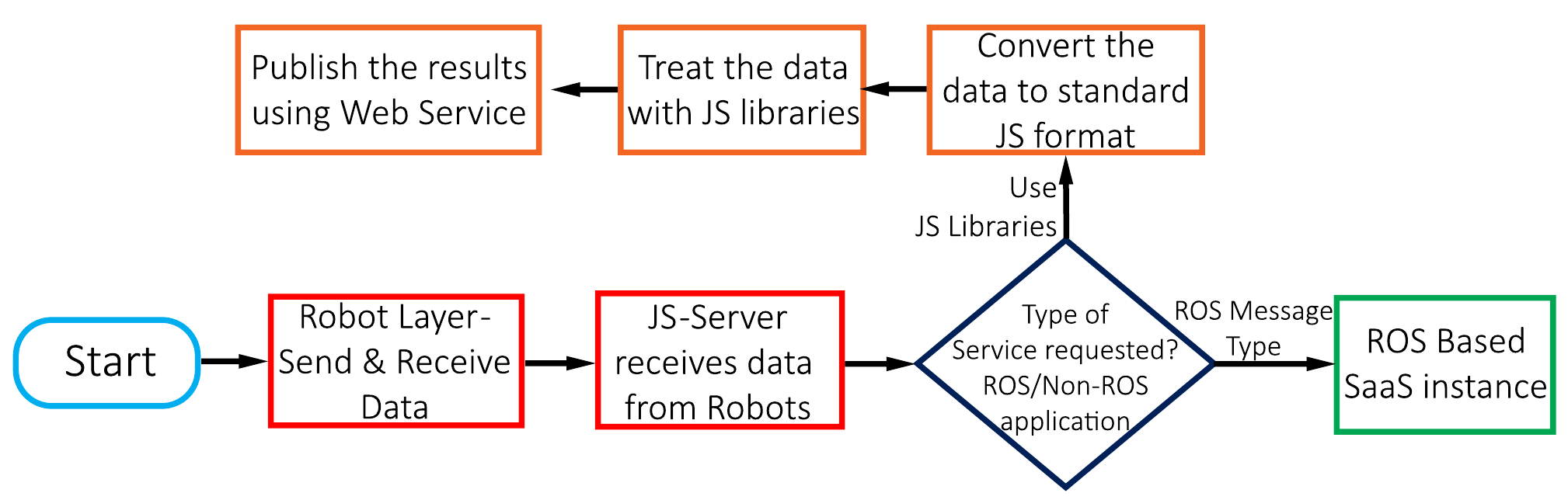}
\caption{Flow chart illustrating the data flow between the robot and Cloud.}
\label{fig:Flow Chart}
\end{figure}

For ROS-based data, the user can choose applications from ROS packages or JS libraries. If the user chooses a JS-based library, the framework decodes the data provided by the robot using appropriate JS libraries. We use inherent JS programming tools like $DataType.type()$ to classify the type of data and then decode it into an appropriate form. For ROS image formats, we use the $Canvas$ library to the convert it to Base64 image format. Figure. \ref{fig:Flow Chart} illustrates the data flow on the cloud.

\subsubsection{JavaScript Libraries}
For the past five years, JS has consistently ranked among the top ten programming languages. It is used everywhere: in web browsers, mobile applications, games, the Internet of Things, robotics, and more. Due to the high usage of JS, the ecosystem around it is growing at a fast pace. Node Package Manager (NPM) is a popular package manager for JS, with more than 35,000 open-source packages.  We intend to use some of this vast set of publically-available libraries in our proposed framework, including the popular machine learning library TensorFlow, which has recently been introduced to the JS ecosystem. To demonstrate the working of the architecture, we used TensorFlow.js for object recognition.

\section{Application and Evaluation}
\label{sec:application}
We conducted an experiment to demonstrate the working of the architecture and also to evaluate various metrics. This experiment used two robots, a Clearpath Jackal Unmanned ground vehicle (UGV) and an iRobot Roomba powered by ODROID-XU4. The hardware configurations of the devices are summarized in Table \ref{tab:spec}.

\begin{center}
\begin{table}[!h]
\caption{Hardware Configuration of the Robots}
\label{tab:spec}
\begin{tabular}{|p{0.6cm}|p{3.4cm}|p{3.4cm}|}
\hline
             & \multicolumn{1}{c|}{Jackal UGV} & \multicolumn{1}{c|}{ODROID-XU4} \\ \hline
CPU &      Intel core i3-4330T SR180 @ 2.40 GHz  & Samsung Exynos5422 A7 Octa-core \\ \hline
RAM &                     2 GB                &                       2 GB                 \\ \hline
\end{tabular}
\end{table}
\end{center}

\subsection{Offloading ROS applications to the cloud}
To validate the proposed architecture, a computationally intense application is an ideal choice. One such application is gmapping, it continuously takes various sensor inputs such as transforms and laser-scans, and it provides output in the form of a map and entropy. The gmapping package is an ROS wrapper for Openslam’s gmapping \cite{Balaguer_towardsquantitative} that is used to create a map, while the robot navigates through the environment. The gmapping package provides Simultaneous Localization and Mapping (SLAM) using the laser input provided by the robot. The Jackal robot used for the experiment was equipped with Velodyne laser scanner, and the input from this laser scanner was used to generate the map of the environment. 

For this experiment, we implemented the ROS package gmapping over the cloud and also on the robot. The architecture subscribed to the topics from the robot and executed the gmapping application on the cloud. We evaluated the performance impact using two metrics: CPU utilization and latency.

\begin{figure}[!t]
\vspace{2mm}
\centering
\includegraphics[width=0.44\textwidth]{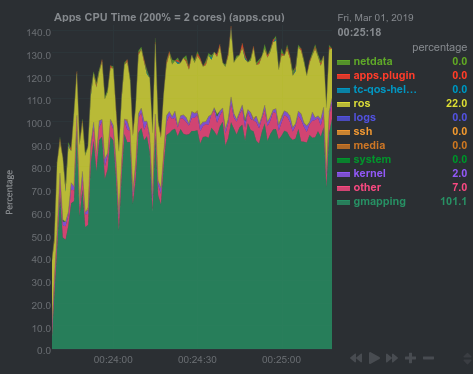}
\caption{CPU utilization with gmapping deployed on the robot (Jackal). On balance, gmapping fully consumes one processor core.}
\label{fig:CPU with GMapping}
\end{figure}

\begin{figure}[!t]
\centering
\includegraphics[width=0.44\textwidth]{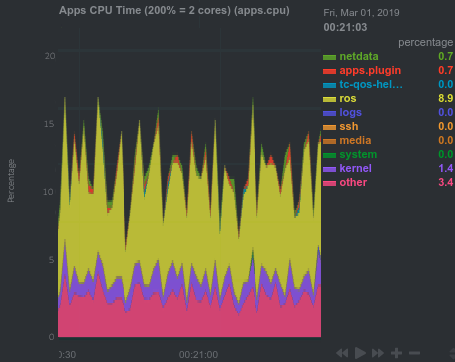}
\caption{CPU utilization with gmapping deployed on the cloud.  Demand on the robot CPU is significantly decreased.}
\label{fig:CPU without GMapping}
\end{figure}

\subsubsection{CPU Utilization}
Execution of the ROS gmapping package onboard the Jackal was compared against the cloud architecture to observe differences in CPU utilization. The Jackal robot has two cores CPU and each core has a capacity of 100 percent; hence, the total CPU capacity is represented as 200 percent. CPU utilization was measured using the open-source tool netdata \cite{NetdataG24:online}, which is a real-time performance monitoring tool for Linux-based systems. When offloading applications, the architecture subscribed to the topics published by the robot and executed all back-end computation related to ROS packages on the cloud. By using the architecture, no load associated with ROS packages was incurred to the onboard CPU, hence we observed a dramatic decrease in CPU utilization. We were able to demonstrate a decrease in CPU utilization by offloading gmapping computation on to the proposed architecture, CPU utilization of the robot decreased by an average of ten-fold compared to running the package on the robot. Fig. \ref{fig:CPU with GMapping} and Fig. \ref{fig:CPU without GMapping} show the difference in the CPU utilization on the robot. The applications that were mainly responsible for the reduction of CPU usage are ROS-Master and gmapping. The ROS-Master consumed an additional 20 to 25 percent of CPU with gmapping running on the robot, whereas with the proposed architecture, the ROS-Master consumed less than 10 percent of CPU. The gmapping application was a computationally intense application for robot navigation. When gmapping was deployed on the robot, we observed an average CPU consumption of 85 percent by the gmapping application, whereas with the architecture there was zero associated computational load on the robot CPU. 


\subsubsection{Latency}
Latency is the term used to describe any kind of delay that occurs during data communication over a network. As the Smart Cloud architecture requires an exchange of information between a robot and a cloud service provider, some latency exists between robot requests and cloud service responses. For this experiment, we used an Amazon Web Services (AWS) server located in North Virginia. 

To measure the time delay between robot requests and cloud service responses, we implemented a ROS service to exchange information and recorded the message timestamps. On average, we observed a time delay of around 35 milliseconds, of which an average of 32 milliseconds was associated with AWS data round trip time. Thus, the time delay contributed by the framework was approximately three milliseconds for processing the application on the cloud. 

\subsection{Object Detection using Tensorflow JS library with Odroid (non-ROS)}
\begin{figure}[!t]
\vspace{2mm}
\centering
\includegraphics[width=0.4\textwidth]{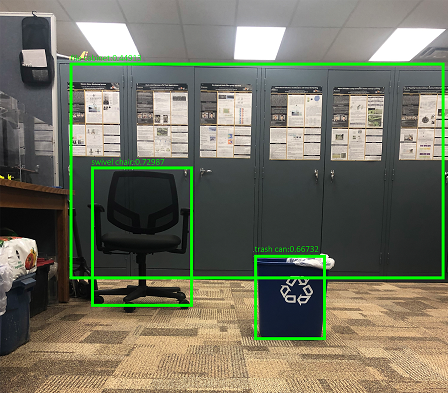}
\caption{Object (i.e. cabinet, swivel chair, and trash can) detected using TensorFlowJS.}
\label{fig:Object Detection}
\end{figure}

In this experiment, we streamed a video from a Roomba equipped with an ODROID-XU4 computer to the architecture. We used Aruco markers along with OpenCV for Roomba to follow Jackal robot. The architecture then identified objects in the video stream using the TensorFlow.js library with an ImageNet dataset and generated XML-based web services to report the results (Fig. \ref{fig:Object Detection}). The following XML file is the web service output generated for Fig. \ref{fig:Object Detection}:\\
\vspace{-5pt}
\begin{Verbatim}[fontsize=\footnotesize, frame=single]
<?xml version="1.0"?>
<Response>
  <Message>
      <MessageID>1</MessageID>
      <ReferenceID></ReferenceID>
      <Result>
          <Class>Trash Can</Class>
           <Probability>0.66</Probability>
      </Result>
      <Result>
         <Class>Swivel Chair</Class>
         <Probability>0.72</Probability>
      </Result>
      <Result>
         <Class>File Cabinet</Class>
         <Probability>0.44</Probability>
      </Result>
  </Message> 
</Response>
\end{Verbatim}

Between the robot publishing a frame and the resultant web service feedback, we observed an average time delay of 34 milliseconds.
The same functionality can be implemented for ROS Image messages by converting the image to base64 format. The following lines of code can be used to convert a ROS Image to JPEG format:\\
\vspace{-5pt}
\begin{Verbatim}[fontsize=\footnotesize, frame=single]
var imgResponse = new Image();
var byteCharacters = atob(message.data);
var abc = "data:image/jpeg;base64,"+byteCharacters;  
imgResponse.src = abc; 
\end{Verbatim}

\subsection{Application scenario using a heterogeneous multi-robot}

\begin{figure*}
\vspace{2mm}
\centering
\includegraphics[width=0.80\textwidth]{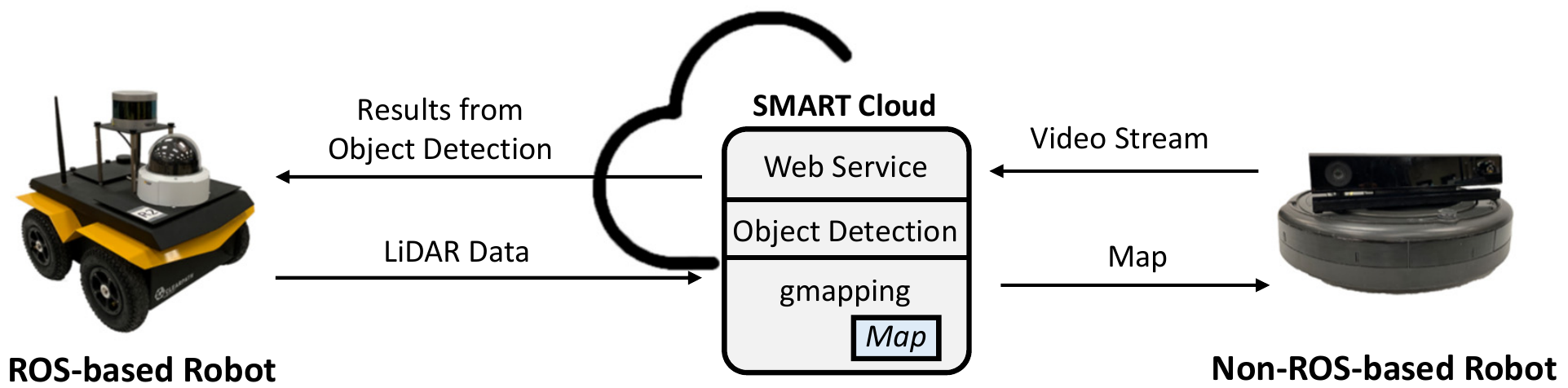}
\caption{ Heterogeneous multi-robot setup.}
\label{fig:Heterogenious}
\end{figure*}
\begin{figure}[!t]
\centering
\includegraphics[width=0.48\textwidth]{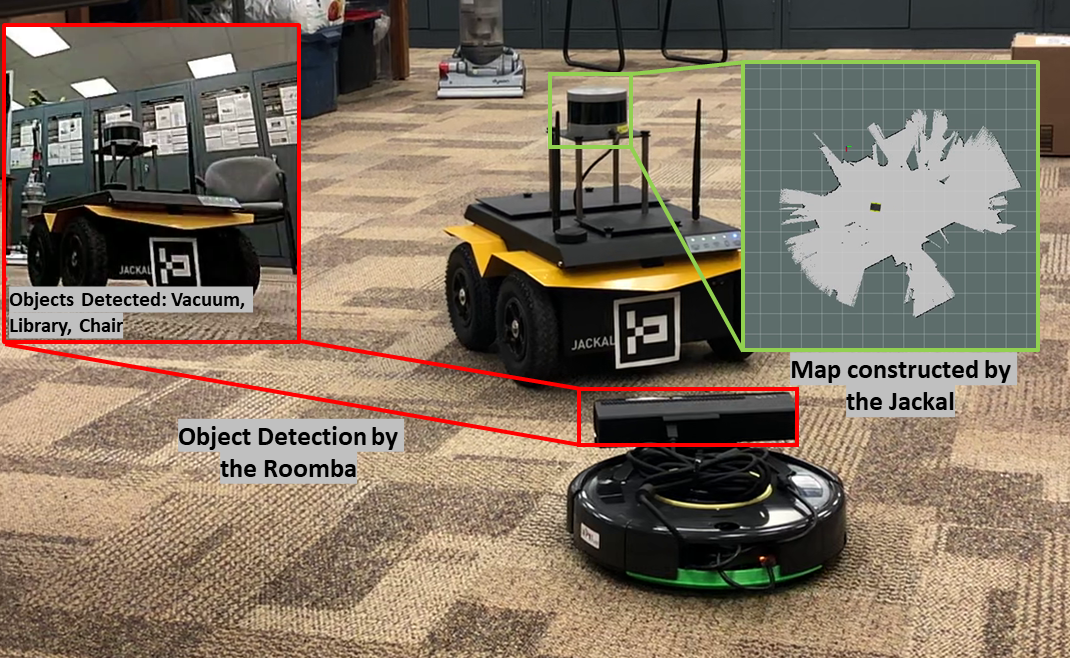}
\caption{Heterogeneous multi-robot application with Jackal and Roomba. Jackal generates a map of the environment and Roomba detects objects in the environment.}
\label{fig:Hetroapplication}
\end{figure}

In this section, we demonstrated a scenario applying a heterogeneous multi-robot system to a collaborative search and rescue operation that was implemented using the proposed architecture. Figure. \ref{fig:Heterogenious} illustrates this heterogeneous system setup. 

The multi-robot system consisted of a non-ROS-based iRobot Roomba equipped with a Microsoft Kinect Camera and a ROS-based Clearpath Jackal equipped with Velodyne 3D LiDAR. The goal of the search and rescue operation was to generate a map and find an object of interest within the map. Figure. \ref{fig:Hetroapplication} shows the heterogeneous multi-robot collaboration between the Roomba and the Jackal. The Jackal robot used the LiDAR data and generated a map of the environment. The Roomba was responsible for object detection in the generated map. The data from the Microsoft Kinect camera on the Roomba was continuously streamed to the cloud service using wireless protocol, and TensorFlow.js was used to detect objects in the video stream. The object detection results were continuously streamed back using the web service. Meanwhile, the ROS-based Jackal subscribed to the web service results and generated the map until the Roomba found the object of interest.  

Therefore, we successfully demonstrated the bi-directional subscription through web services and the other novel architecture features through this experiment. A video of the experiment is available for reference at: \\ {\textcolor{blue}{ \url{https://youtu.be/zImysVWLlFs}}}.

\section{Conclusion}
\label{sec:conclusion}
In the paper, we present the \textit{Smart Cloud} architecture, which is the first of its kind to incorporate JavaScript-based libraries for running diverse robotic applications related to machine learning and more. \textit{Smart Cloud} also leverages the resources provided by cloud service providers for use with robotic applications. The architecture can be used with heterogeneous and homogeneous multi-robot systems as well as single-robot systems. We additionally demonstrated the working of ROS and non-ROS based robot systems with the architecture and the incorporation of JS libraries for robotic applications. We measured the performance of the architecture in terms of onboard CPU usage, and latency. We were able to show significant reduction in onboard CPU usage and achieved an average latency of 35 milliseconds.

Our future work will focus on the development of tools and mechanisms to lower latency even further. We are also working on designing an offloading schema that will facilitate the dynamic offloading of applications based on application memory requirements and criticality. We are also developing a pipeline of tools to improve overall performance of the architecture in terms of metrics such as latency, scalability, interoperability, availability, and security.



\bibliographystyle{IEEEtran}
\bibliography{IEEEabrv,manoj}

\end{document}